\pdfoutput=1

\documentclass[11pt]{article}

\usepackage[]{acl}

\usepackage{times}
\usepackage{latexsym}
\usepackage{color,graphicx}
\usepackage{subcaption}

\usepackage{tikz}
\usepackage{tikz-dependency}
\usetikzlibrary{%
  shapes,%
  arrows,%
  positioning,%
  calc,%
  automata%
}
\definecolor{pf7}{RGB}{166, 118, 29}

\usepackage[T1]{fontenc}

\usepackage[utf8]{inputenc}

\usepackage{microtype}

\usepackage{graphicx}

\usepackage{amsmath}
\usepackage{amsfonts}
\usepackage{amssymb}

\usepackage{multirow, multicol}
\usepackage{pgfplots}

\usepackage{enumitem}
\usepackage{booktabs}       

\title{Modeling Label Correlations for Second-Order Semantic Dependency Parsing with Mean-Field Inference}

\author{Songlin Yang, Kewei Tu\\
  School of Information Science and Technology, ShanghaiTech University \\
    {\tt \{yangsl,tukw\}@shanghaitech.edu.cn}\\
 }

\begin{document}
\maketitle
\begin{abstract}
Second-order semantic parsing with end-to-end mean-field inference has been shown good performance. 
In this work we aim to improve this method by modeling label correlations between adjacent arcs. However, direct modeling leads to memory explosion because second-order score tensors have sizes of $O(n^3L^2)$ ($n$ is the sentence length and $L$ is the number of labels), which is not affordable. To tackle this computational challenge, we leverage tensor decomposition techniques, and interestingly, we show that the large second-order score tensors have no need to be materialized during mean-field inference, thereby reducing the computational complexity from cubic to quadratic.
We conduct experiments on SemEval 2015 Task 18 English datasets, showing the effectiveness of modeling label correlations. Our code is publicly available at \url{https://github.com/sustcsonglin/mean-field-dep-parsing}. 

\end{abstract}

\section{Introduction} 
\begin{figure}[tb!]
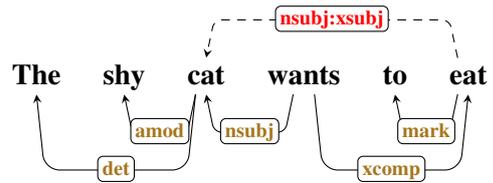
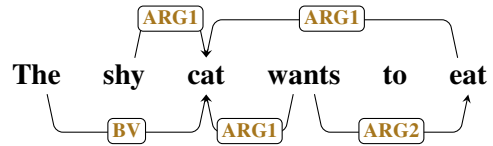
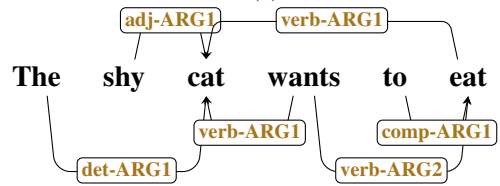
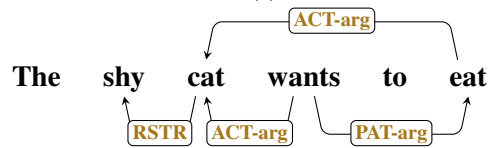

 \begin{subfigure}{\linewidth}
\begin{dependency}
        \begin{deptext}[column sep=1em, row sep=1em]
            \textbf{The} \& \textbf{shy} \& \textbf{cat} \& \textbf{wants} \& \textbf{to} \& \textbf{eat} \\
        \end{deptext}
        \depedge[edge below,edge height=1cm]{3}{1}{\bf\textcolor{pf7}{det}}
        \depedge[edge below,edge height=0.5cm]{3}{2}{\bf\textcolor{pf7}{amod}}
        \depedge[edge below,edge height=0.5cm]{4}{3}{\bf\textcolor{pf7}{nsubj}}
        \depedge[edge below,edge height=1cm]{4}{6}{\bf\textcolor{pf7}{xcomp}}
        \depedge[edge below,edge height=0.5cm]{6}{5}{\bf\textcolor{pf7}{mark}}
        \depedge[edge above,edge height=0.5cm, dashed]{6}{3}{\bf\textcolor{red}{nsubj:xsubj}}
    \end{dependency}
 \subcaption{Stanford typed dependencies with enhanced dependencies dashed and colored in red.}
 \end{subfigure}
  \begin{subfigure}{\linewidth}
\begin{dependency}
        \begin{deptext}[column sep=1em, row sep=1em]
            \textbf{The} \& \textbf{shy} \& \textbf{cat} \& \textbf{wants} \& \textbf{to} \& \textbf{eat} \\
        \end{deptext}
        \depedge[edge below, edge height=0.5cm]{1}{3}{\bf\textcolor{pf7}{BV}}
        \depedge[edge above, edge height=0.5cm]{2}{3}{\bf\textcolor{pf7}{ARG1}}
        \depedge[edge below, edge height=0.5cm]{4}{3}{\bf\textcolor{pf7}{ARG1}}
        \depedge[edge above, edge height=0.5cm]{6}{3}{\bf\textcolor{pf7}{ARG1}}
        \depedge[edge below, edge height=0.5cm]{4}{6}{\bf\textcolor{pf7}{ARG2}}
    \end{dependency}
 \subcaption{DM}
 \end{subfigure}
 \begin{subfigure}{\linewidth}

\begin{dependency}
        \begin{deptext}[column sep=1em, row sep=1em]
            \textbf{The} \& \textbf{shy} \& \textbf{cat} \& \textbf{wants} \& \textbf{to} \& \textbf{eat} \\
        \end{deptext}
        \depedge[edge below, edge height=1cm]{1}{3}{\bf\textcolor{pf7}{det-ARG1}}
        \depedge[edge above, edge height=0.5cm]{2}{3}{\bf\textcolor{pf7}{adj-ARG1}}
        \depedge[edge below, edge height=0.5cm]{4}{3}{\bf\textcolor{pf7}{verb-ARG1}}
        \depedge[edge height=0.5cm]{6}{3}{\bf\textcolor{pf7}{verb-ARG1}}
        \depedge[edge below, edge height=1cm]{4}{6}{\bf\textcolor{pf7}{verb-ARG2}}
        \depedge[edge below, edge height=0.5cm]{5}{6}{\bf\textcolor{pf7}{comp-ARG1}}
    \end{dependency}
 \subcaption{PAS}
 \end{subfigure}
 \begin{subfigure}{\linewidth}

\begin{dependency}
        \begin{deptext}[column sep=1em, row sep=1em]
            \textbf{The} \& \textbf{shy} \& \textbf{cat} \& \textbf{wants} \& \textbf{to} \& \textbf{eat} \\
        \end{deptext}
        \depedge[edge below,edge height=0.5cm]{3}{2}{\bf\textcolor{pf7}{RSTR}}
        \depedge[edge below,edge height=0.5cm]{4}{3}{\bf\textcolor{pf7}{ACT-arg}}
        \depedge[edge height=0.5cm, edge below]{4}{6}{\bf\textcolor{pf7}{PAT-arg}}
        \depedge[edge height=0.5cm]{6}{3}{\bf\textcolor{pf7}{ACT-arg}}
    \end{dependency} \subcaption{PSD}
 \end{subfigure}
\caption{Comparison between syntactic and semantic dependency schemes. Roots are omitted for brevity.}
\label{fig:dep}
\end{figure}

Syntactic dependency parsing has been well-studied and widely used in natural language processing. It provides approximation to the semantic relationship between dependent and head words, which is useful in many downstream natural language understanding tasks. However, the underlying tree-structured representation of syntactic dependency parsing limits the number and types of relationships that can be captured. For example, in Fig.\ref{fig:dep}(a), the word $\textit{cat}$ is the subject of both $\textit{wants}$ and $\textit{eat}$, and both of these two relationships could benefit downstream tasks. However, within a dependency tree, each word has exactly one head, thus the two relationships cannot be modeled simultaneously. To enrich the representation power, semantic dependency parsing \cite{oepen-etal-2014-semeval} aims to produce a labelled directed acyclic graph (DAG) for a given sentence, whereby each word can be attached to zero, one, or multiple heads. Fig.\ref{fig:dep}(b-d) show three different semantic formalisms: DELPH-IN MRS (DM) \cite{flickinger2012deepbank}, Predicate-Argument Structure (PAS) \cite{DBLP:conf/coling/MiyaoT04}, and Prague Semantic Dependencies (PSD) \cite{hajic-etal-2012-announcing}, respectively.

\begin{figure}[tb!]
    \centering
    \includegraphics[width=1\linewidth]{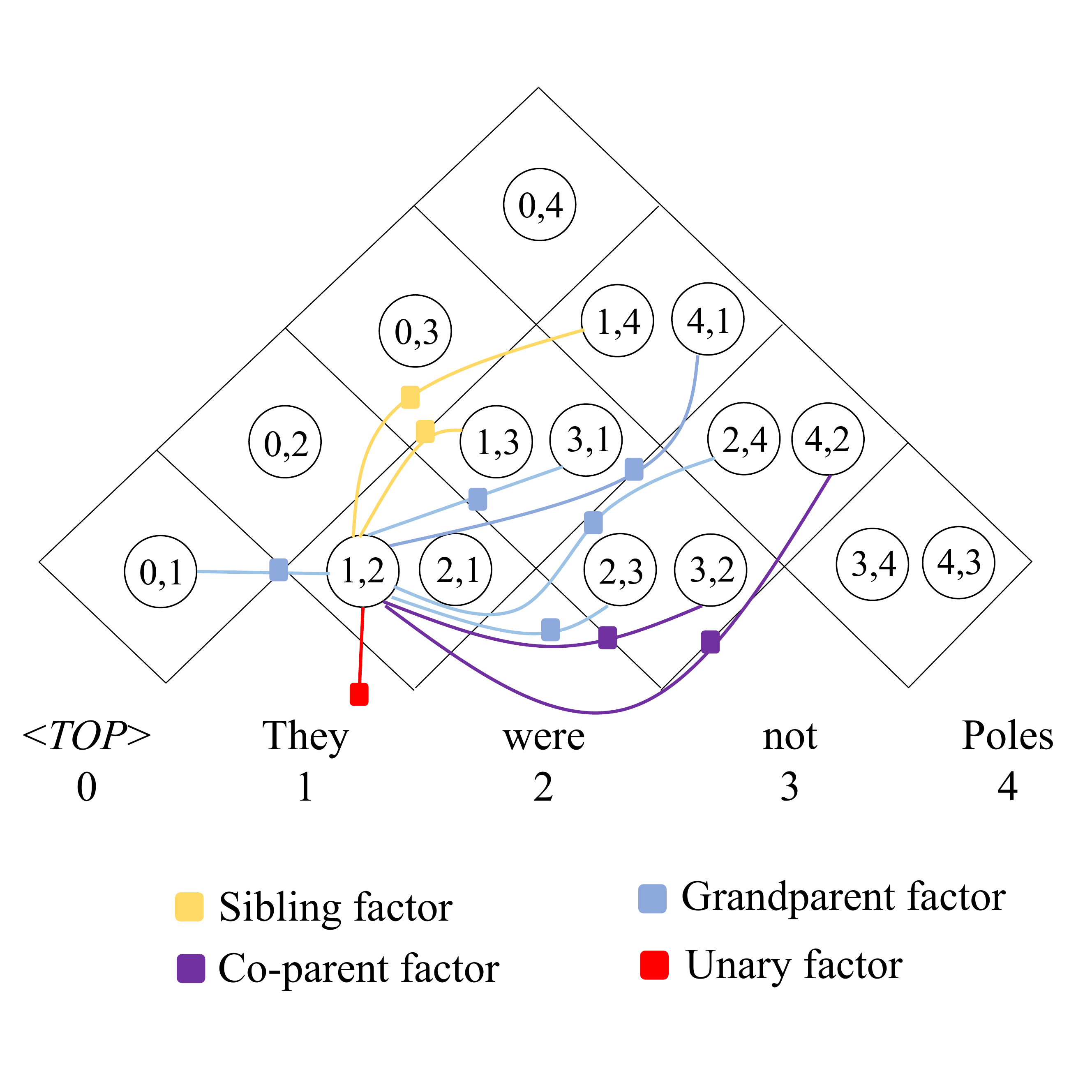}
    \caption{Factor graph representation of second-order dependency parsing. The figure comes from \citet{wang-etal-2019-second}.}
    \label{fig:crf}
\end{figure}

Graph-based parsers achieve great success in dependency parsing. Higher-order parsers take into accounts interaction between multiple dependency arcs, thus are more powerful than first-order parsers. However, exact higher-order parsing is intractable except for projective tree parsing, making approximate decoding a necessity. Both syntactic and semantic dependency parsing can be formulated as an inference problem on Markov Random Fields (MRFs) \cite{smith-eisner-2008-dependency,wang-etal-2019-second} where dependency arcs are nodes, and higher-order parts (e.g., sibling and grandparent relationships) are factors, as shown in Fig. \ref{fig:crf}. In this sense, approximate inference algorithms of graphical models can be used for approximate parsing. Previous researchers have used
 loopy belief propagation \cite{smith-eisner-2008-dependency, gormley-etal-2015-approximation, wang-etal-2019-second}, mean-field inference \cite{wang-etal-2019-second, wang-tu-2020-second}, and alternating directions dual decomposition (AD$^3$) \cite{martins-etal-2011-dual, martins-etal-2013-turning,martins-almeida-2014-priberam, fonseca-martins-2020-revisiting} for higher-order syntactic or semantic dependency parsing. The iterative inference steps of these inference algorithms are fully differentiable and can be unrolled as recurrent neural network layers for end-to-end learning \cite{Domke2011ParameterLW,  DBLP:conf/iccv/0001JRVSDHT15, gormley-etal-2015-approximation,wang-etal-2019-second,wang-tu-2020-second, DBLP:conf/icml/NiculaeM20}. 

In this work, we aim to improve second-order semantic dependency parsing with end-to-end mean-field inference \cite{wang-etal-2019-second} by modeling label correlations between adjacent arcs. Intuitively, it is beneficial to model label correlations. For example, in Fig.\ref{fig:dep}(b), the arc $\textit{wants}\rightarrow\textit{cat}$ labeled with \texttt{verb-AGR1} and the arc $\textit{wants}\rightarrow \textit{eat}$ labeled with \texttt{verb-ARG2} are siblings. If the second-order sibling scores take labels into account, it is less likely to mistakenly classify both arcs as \texttt{verb-ARG1} or \texttt{verb-ARG2}.  Despite the obvious advantage of modeling label correlations, to the best of our knowledge, most if not all of existing second-order parsers only model \textbf{unlabeled} adjacent arcs, instead of the \textbf{labeled} ones. A possible reason is that taking labels into accounts would scale up the size of second-order score tensors from $O(n^3)$ to $O(n^3L^2)$, where $n$ is the sentence length and $L$ is the number of label, thereby leading to memory explosion. 
 To tackle this computational challenge, we apply canonical-polyadic decomposition (CPD) \cite{DBLP:journals/corr/abs-1711-10781} on second-order score tensors, and  interestingly, mean-field inference can be greatly accelerated because the large second-order score tensors have no need to be materialized. Instead, only several smaller tensor multiplication operations are involved, thereby reducing the computational complexity from cubic to quadratic.  Our contribution can be summarized as follows:
\begin{itemize}
    \item We decrease the time and space complexity of second-order parsing with mean-field inference from cubic to quadratic.
    \item We validate the effectiveness of modeling label correlations on SemEval 2015 Task 18 English datasets.
\end{itemize}

\section{Background}

\subsection{Semantic dependency parsing as energy minimization}
Given a sentence $x = x_0, x_1, \cdots x_n$ with $x_0$ being the root token, its labeled dependency graph $y$\footnote{We use $y$ to represent a dependency graph or its indicator tensor alternatively whenever the context is clear.} can be represented as an order-3 indicator tensor $y \in R^{(n+1) \times (n+1) \times L}$ where $y_{ijl} = 1$ iff there is an arc $x_i \rightarrow x_j$ with label $l$ and $y_{i j l} = 0$ otherwise; $L$ the size of label set.
 Dependency parsing can be formulated as an inference problem on MRFs \cite{smith-eisner-2008-dependency}. Each variable node $(i, j)$ stands for a dependency arc $x_i \rightarrow x_j$, and arc correlations such as sibling and grandparent relationships are represented as factors. Fig. \ref{fig:crf} shows an example. Inference with MRFs is known as energy minimization. We can define the negative energy of a \textbf{labeled} dependency graph $y$ as:
 \begin{align}
 &-E(y) =\sum_{ija} s_{ija}^{\text{arc}} y_{ija} + \frac{1}{2} ( \sum_{i j k a b} s^{\text{sib}}_{i j k a b} y_{ija} y_{i k b} \nonumber  \\ 
 &+ \sum_{i j k a b} s^{\text{cop}}_{i j k a b} y_{i j a} y_{k j b} + \sum_{i j k a b} s^{\text{grd}}_{i j k a b} y_{i j a} y_{j k b})
 \label{eq:energy}
\end{align} 
where $s^{\text{arc}} \in R^{(n+1)\times (n+1) \times L}$ is the arc score, $s^{\text{sib}}, s^{\text{cop}}, s^{\text{grd}} \in R^{(n+1) \times (n+1) \times (n+1) \times L}$ are the sibling scores, co-parent scores and grandparent scores, respectively. The label correlations are captured by the second-order scores. For example, $s^{sib}_{ijkab}$ specifies how likely an arc $x_{i} \rightarrow x_{j}$ with label $a$ and another arc $x_{i} \rightarrow x_{k}$ with label $b$ exist simultaneously. Then dependency parsing becomes a discrete energy minimization problem: $y^{opt} = \arg \min_{y \in \mathcal{Y}} E(y)$
where $\mathcal{Y}$ is the set of valid dependency graphs.
Since discrete energy minimization is intractable in general, one often makes local polytope relaxation on the discrete variable and round the final continuous solution into the closest discrete one. We denote $\Delta$ as the local polytope relaxation of $\mathcal{Y}$ where $y \in \Delta$ iff $0\le y_{ijl} \le 1; \sum_{l}y_{ijl}=1$ for all $i,j,l$. Note that the first label is preserved for the empty label \texttt{NULL}. 

 
\subsection{Mean-field inference}
There are many perspectives and understandings of the mean-field inference, and we take the energy minimization view: \citet{le2021regularized} show that mean-field can be formulated as iteratively  using conditional gradient descent to optimize the energy function, but with an additional entropy penalty as follows:
\begin{align}
y^{m+1} &=  \arg\min_{y \in \Delta}   \langle\, F^{m}, y \rangle - H(y) \nonumber \\ 
        &=  \arg\max_{y \in \Delta}  \langle\, -F^{m}, y \rangle + H(y) 
\label{eq:mf}
\end{align} 
where m is the iteration number; $F^m :=\nabla E(y^m)$; $\langle\,\rangle$ is the (Frobenius) inner product; and  $H$ is the entropy:
\[
H(y) = - \sum_{i j l} y_{ijl}\log(y_{i j l}) 
\]
Note that Eq. \ref{eq:mf} is the variational representation of the softmax function, and thus the solution is:
\begin{equation}
y^{m+1}_{ijl} = \frac{\exp(-F^{m}_{ijl})}{\sum_{l^{\prime}}\exp(-F^{m}_{ijl^\prime})}  
\label{eq:maximizer}
\end{equation}
Put Eq. \ref{eq:energy} into $F^m =\nabla E(x^m)$, we have 
\begin{align}
-F_{ija}^m = s^{arc}_{ija} \nonumber &+ \sum_{kb}(s^{sib}_{i j kab} y^{m}_{i kb} \\ &+ s^{cop}_{i j kab} y^{m}_{k jb} + s^{grd}_{i j kab} y^{m}_{j kb})
\label{eq:update}
\end{align}
and we set the initial energy $-F^{0} = s^{\text{arc}}$.  In summary, mean-filed inference has two main steps during iterations, (1) each node aggregates the scores from all neighbors and itself (Eq. \ref{eq:update}) based on the posteriors of the last iteration, and (2) the posteriors are updated by softmax mapping on the aggregated scores(Eq.\ref{eq:mf},\ref{eq:maximizer}).

\subsection{CP decomposition}
Given a tensor $T \in R^{N_1 \times N_2 \cdots \times N_m}$ with $N_k$ possible values in the $k$th dimension for each $k \in \{1\cdots m\}$, canonical-polyadic decomposition (CPD) aims to find a compact representation of $T$ by decomposing it into the sum of outer product of vectors:
\begin{equation}
\mathbf{T}=\sum_{r=1}^{R} \lambda_{r} w_{r 1} \otimes w_{r 2} \otimes \cdots \otimes w_{r m}
\label{eq:cpd}
\end{equation}
where $\lambda_r \in R$ is the scalar coefficient, and can be absorbed into $\{w_{rk}\}$, thus we omit it throughout the paper; $w_{r k} \in R^{N_k}$; $\otimes$ is the outer product. In this way, we only need $R \times (N_1 + N_2 + \cdots + N_k)$ parameters to compactly represent the original tensor $T$ whose size grows exponentially. 

\section{Method}
\begin{figure*}[tb!]
    \centering
    \includegraphics[width=0.8\linewidth]{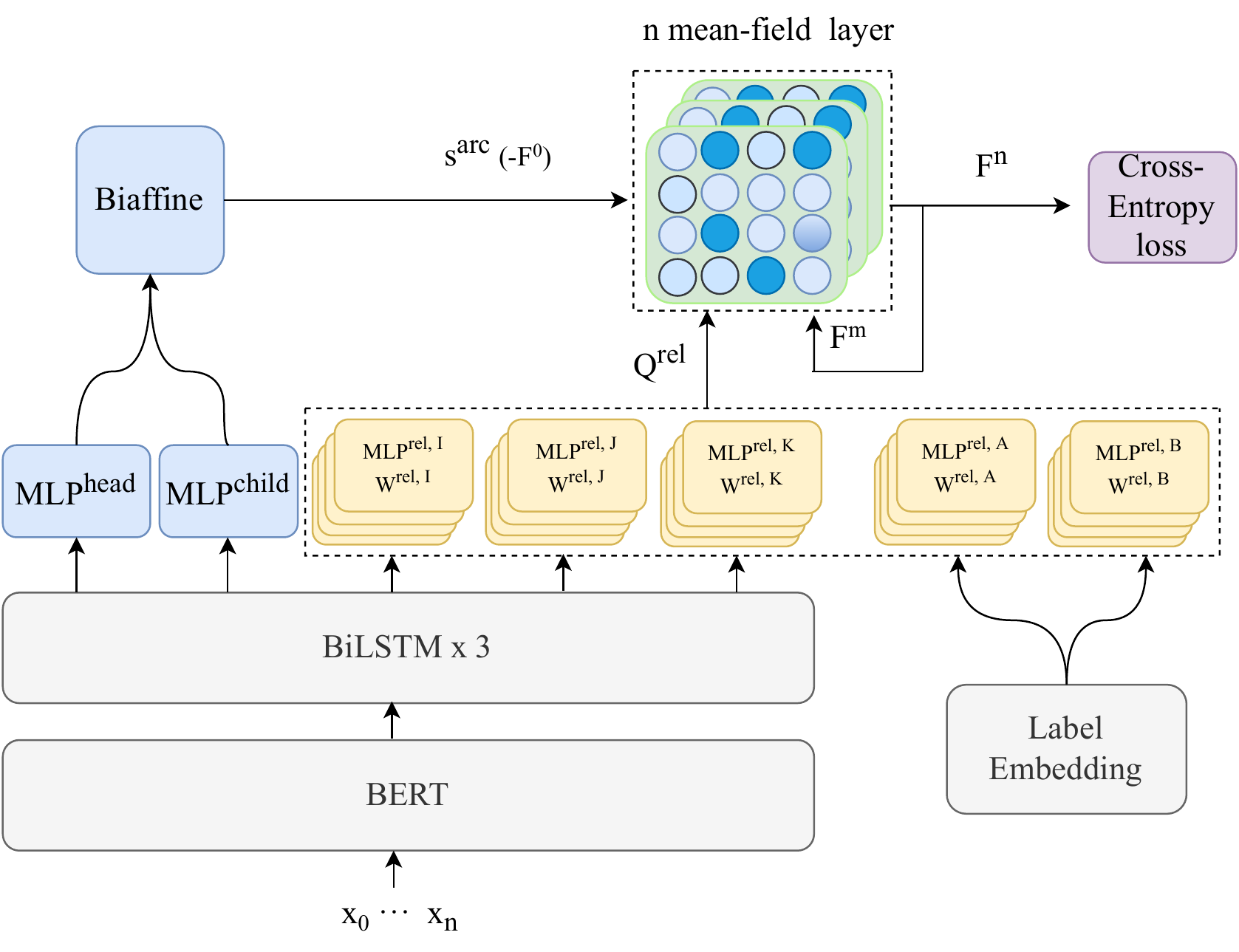}
    \caption{Model architecture.}
    \label{fig:net}
\end{figure*}
\subsection{Speed-up mean-field inference via CPD}
We aim to improve the computational complexity of Eq.\ref{eq:update}. We take sibling scores for example to demonstrate our idea, and other second-order scores can be manipulated in a similar fashion.  
Consider $t$ defined as:
\begin{equation}
t_{ija} := \sum_{k,b} s^{sib}_{ijkab}y_{ikb}^m
\label{eq:term}
\end{equation}
we find it convenient to use the Einstein summation (einsum) notations, and rewrite it as:
\[
t = \operatorname{einsum}(``ijkab, ikb\rightarrow ij", s^{sib}, y^m)
\]
The computational complexity is $O(n^3L^2)$. To reduce it, we apply CPD on the order-5 tensor $s^{sib}$,
\[
s^{sib} = \sum_{r=1}^{R} w_{ri} \otimes w_{rj} \otimes w_{rk} \otimes w_{ra} \otimes w_{rb}
\]
where $w_{ri}, w_{rj}, w_{rk} \in R^{n+1}, w_{ra}, w_{rb} \in R^{L}$. For each $t \in \{i,j,k,a,b\}$, stacking all $w_{rt}$ for all $r$ column-wise, we obtain five matrices $I^{sib}, J^{sib}, K^{sib} \in R^{(n+1)\times R}, A^{sib}, B^{sib} \in R^{L \times R}$.   Then we have
\[
s^{sib}_{ij k a b} = \sum_{r=1}^{R} I^{sib}_{i r} J^{sib}_{j r} K^{sib}_{k r} A^{sib}_{ar} B^{sib}_{b r}
\]
and Eq. \ref{eq:term} can be written as:
\begin{align*}
t_{ija} &= \sum_{kb} \sum_{r=1}^{R}  I^{sib}_{ir} J^{sib}_{jr} K^{sib}_{kr} A^{sib}_{ar} B^{sib}_{br} y_{ikb}^m  \\
 &= \sum_{r=1}^{R}  J^{sib}_{jr} A_{ar}^{sib} \underbrace{\sum_{kb} I^{sib}_{ir}  K^{sib}_{kr} B^{sib}_{br}  y_{i k b}^m}_{\texttt{Term1}}
\end{align*}
The key insight here is that we can change the order of summation, cache \texttt{Term1} so that it can be reused for arbitrary $j, a$, thereby reducing computational complexity. The above equation can be written as the following einsum form,
\begin{align*}
t^\prime&=\operatorname{einsum}(``ir, kr, br, ikb \rightarrow ir", \\  &\quad\quad\quad\quad\quad I^{sib}, K^{sib}, B^{sib}, y^m)\\
t &=\operatorname{einsum}(``ir,  jr, ar \rightarrow ija", t^\prime, J^{sib}, A^{sib}) 
\end{align*}
which reduces the computational complexity from cubic to quadratic, i.e., $O(n^2LR)$ with $R \ll nL$.
Notably, $s^{sib}$ is never materialized during iterations.

\subsection{Neural scoring}
We use neural networks to compute $s^{arc}, Q^{rel}$ where $Q\in \{I, J, K, A, B\}, rel \in \{sib, cop, grd\}$. Fig. \ref{fig:net} depicts our model architecture. We feed the sentence $x=x_0\cdots x_n$ ($x_0$ is the root token) into BERT \cite{devlin-etal-2019-bert} to obtain contextualized word embeddings:
\[c_{i} = \operatorname{BERT}(x_i)
\]
and we apply mean-pooling to the last layer of BERT to obtain word-level embedding. We concatenate $c$ with POS tag and Lemma embeddings:
\[
 e_{i} =  c_{i} \oplus e_{i}^{\text{pos}} \oplus e_{i}^{\text{lemma}}
\]
and feed $e_{0} \cdots e_{n}$ into a three-layer bidirectional LSTM \cite{hochreiter1997long} (BiLSTM):
 \begin{align}
  & \dots, (\overrightarrow{b_i}, \overleftarrow{b_i}), \dots = \operatorname{BiLSTM}([\dots, e_i, \dots]) \nonumber 
\end{align}
Then we use deep biaffine attention \cite{DBLP:conf/iclr/DozatM17} to compute $s^{arc}$:
\begin{align*}
&e^{head/child}_i = \operatorname{MLP}^{head/child}([\overrightarrow{b_i}; \overleftarrow{b_{i}}]) \\
&s^{arc}_{i, j, l} = [e^{head}_i; 1]^T W^l [e^{child}_j;1])
\end{align*}
where $W^l \in R^{(k+1) \times (k+1)}$ is trainable parameter. For $Q^{rel}$, denote the label embedding matrix as $P$, we first obtain type-specific representations:
\begin{align*}
&e^{rel, I/J/K}_{i} = \operatorname{MLP}^{rel, I/J/K}([\overrightarrow{b_i}; \overleftarrow{b_{i}}])\\
&e^{rel, A/B}_{i} = \operatorname{MLP}^{rel, A/B}(P_i) \\
\end{align*}
Then we apply affine transformations to compute $Q^{rel}$:
\begin{align*}
     Q^{rel}_i = [e^{rel, Q}_i;1] W^{rel, Q} 
\end{align*} 
where $W^{rel, Q} \in R^{(k+1) \times R}$.


\subsection{Loss and parsing}
After running $n$ iterations of mean-field inference, we obtain the final energy $F^n$ (Eq. \ref{eq:update}). Let $t_{ij}$ denote the index of the label of arc $x_i \rightarrow x_j$. If there is no such arc in the gold dependency graph, we set $t_{ij}$ to the index of the \texttt{NULL} label, i.e.,  $t_{ij} = 0$.  Then we use cross-entropy to define the loss $L$:
\[
L = -\sum_{i, j}\log \frac{\exp(-F^{n}_{ijt_{ij}})}{\sum_{l^{\prime}} \exp(-F^{n}_{ijl^{\prime}})}
\]
Since mean-field inference is fully differentiable, we use automatic differentiation to update parameters.
For parsing, let
$y^{\star}_{ij} = \arg\max_l -F^n_{ijl}$.  If $y^{\star}_{ij} = 0$, i.e., the predicted label is \texttt{NULL}, then the arc $x_i \rightarrow x_j$ does not exist, otherwise we add it to the final predicted semantic graph.

\section{Experiments}

\subsection{Setup}
We conduct experiments on the SemEval 2015 Task 18 English
datasets \cite{oepen-etal-2015-semeval}. Sentences are annotated with three formalism: DM, PAS, and PSD.
We use the same data splitting as previous works \cite{martins-almeida-2014-priberam, du-etal-2015-peking} with 33,964 sentences in the training set, 1,692 sentences in the development set, 1,410 sentences in the in-domain (ID) test set and 1,849 sentences in the out-of-domain (OOD) test set from the Brown Corpus \cite{francis1982frequency}. We use POS tags and lemmas as additional features, use ``bert-base-cased'' as contextual word embedding.  We report the labeled F-measure scores (LF1) in the ID and OOD test sets for each formalism. The reported results are averaged over three runs with different random seeds. In each run, we select the best model based on the performance on the development set. 

\subsection{Hyper-parameters}
The hyper-parameters are summarized in Table \ref{tab:hyper}. Besides, the maximum training epoch is set to 30 for DM; 20 for PAS and PSD. 
\begin{table}
\begin{footnotesize}
\centering
\begin{tabular}{@{\hskip 0pt}lc@{\hskip 0pt}}
\hline
\textbf{Architecture hyper-parameters} & \\
\hline
BERT embedding dimension & 768\\
POS/Lemma embedding dimension & 100\\
Embeddings dropout & 0.33 \\
BiLSTM encoder layers & 3 \\
BiLSTM encoder size & 1000 \\
BiLSTM layers dropout & 0.33 \\
MLP layers & 1 \\
MLP activation function & LeakyReLU \\
MLP layers dropout  & 0.33\\
MLP dimension & 300 \\ 
Rank dimension & 300\\
Mean-field inference iterations (training) & 2\\
Mean-field inference iterations (testing) & 10 \\
\hline
\textbf{Hyper-parameters regarding training} &\\
\hline
BERT learning rate & 5e-5 \\
Other learning rate & 2.5e-3\\
Optimizer & AdamW \\ 
Scheduler & linear warmup \\
Warmup rate & 0.5 \\ 
Gradient clipping & 5.0 \\
Tokens per batch & 3000 \\
Maximum training sentence length & 150 \\ 
\hline
\multicolumn{1}{c}{}\\
\end{tabular}
\centering
\setlength{\abovecaptionskip}{4pt}
\caption{Summary of hyper-parameters.}
\label{tab:hyper}
\end{footnotesize}
\end{table}




\section{Main Result}

\begin{table*}[tb!]
\centering
\begin{tabular}{@{\hskip 0.5pt}lcccccccc@{\hskip 0.5pt}}
& \multicolumn{2}{c}{DM}
& \multicolumn{2}{c}{PAS}
& \multicolumn{2}{c}{PSD}
& \multicolumn{2}{c}{Avg}
\\
Parser & ID & OOD & ID & OOD & ID & OOD & ID & OOD \\
\hline
\citet{DBLP:conf/iclr/DozatM17} \scriptsize{\textbf{+char+lemma}}   & 93.7 & 88.9 & 93.9 & 90.6 & 81.0 & 79.4 & 89.5 & 86.3 \\
\citet{kurita-sogaard-2019-multi} \scriptsize{\textbf{+lemma}}   & 92.0 & 87.2 & 92.8 & 88.8 & 79.3 & 77.7 & 88.0 & 84.6 \\
\citet{wang-etal-2019-second} (MF) \scriptsize{\textbf{+char+lemma}}   & 94.0 & 89.7 & 94.1 & 91.3 & 81.4 & 79.6 & 89.8 & 86.9 \\
\citet{wang-etal-2019-second} (LBP) \scriptsize{\textbf{+char+lemma}}   & 93.9 & 89.5 & 94.2 & 91.3 & 81.4 & 79.5 & 89.8 & 86.8 \\
Pointer \scriptsize{\textbf{+char+lemma}} & 93.9 & 89.6 & 94.2 & 91.2 & 81.8 & 79.8 & 90.0 & 86.9 \\
\hline
\citet{zhang-etal-2019-broad} \scriptsize{+char\textbf{+BERT$_{\tt LARGE}$}}   & 92.2 & 87.1 & - & - & - & - & - & - \\
\citet{lindemann-etal-2019-compositional} \scriptsize{\textbf{+BERT$_{\tt BASE}$}} & 94.1 & 90.5 & 94.7 & 92.8 & 82.1 & 81.6 & 90.3 & 88.3 \\
\citet{lindemann-etal-2020-fast} \scriptsize{\textbf{+BERT$_{\tt LARGE}$}} & 93.9 & 90.4 & 94.7 & 92.7 & 81.9 & 81.6 & 90.2 & 88.2  \\ 
\citet{DBLP:conf/flairs/HeC20} \scriptsize{+lemma\textbf{+Flair+BERT$_{\tt BASE}$}}   & 94.6 & 90.8 & \textbf{96.1} & 94.4 & \textbf{86.8} & 79.5 & 92.5 & 88.2 \\
Pointer \scriptsize{+char+lemma\textbf{+BERT$_{\tt BASE}$}} & 94.4  & 91.0 & 95.1 & 93.4 & 82.6 & 82.0 & 90.7 & 88.8  \\
Ours \scriptsize{+lemma\textbf{+BERT$_{\tt BASE}$}}   & \textbf{95.0}  & \textbf{91.8} & 95.4 & 93.5 & 82.7 & \textbf{82.2} &  91.0 & \textbf{89.2} \\
\quad w/o label correlation & 94.8 &  91.7 &  94.6 & 93.0 & 82.4 & 81.6 & 90.6 & 88.8  \\
\hline
\multicolumn{1}{c}{}\\
\end{tabular}
\centering
\setlength{\abovecaptionskip}{4pt}
\caption{Labeled F1 scores on three formalisms of SemEval 2015 Task 18. $+\mathit{char}$ and $+\mathit{lemma}$ means using character and lemma embeddings.
 Pointer: \citet{fernandez-gonzalez-gomez-rodriguez-2020-transition}. 
}
\label{tab:sdp_result}
\end{table*}
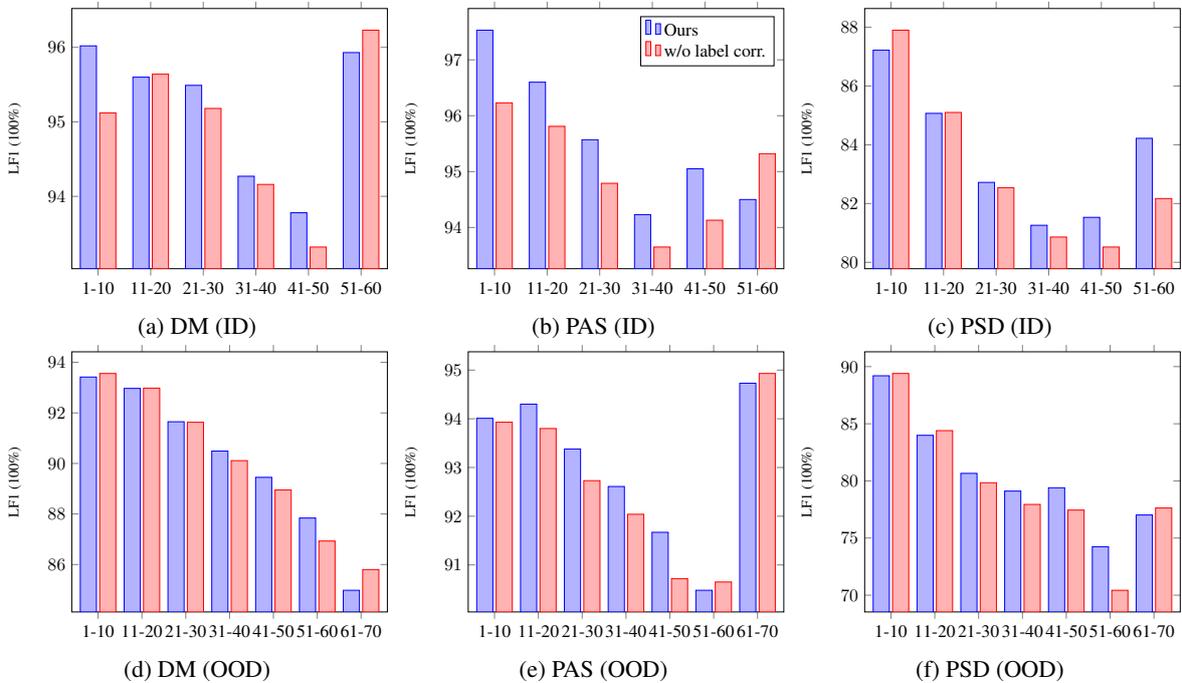
\begin{figure*}[tb!]
\centering
	\begin{subfigure}[t]{0.32\linewidth}
		 \resizebox{\textwidth}{!}{%
		 
		 \begin{tikzpicture}
  \begin{axis}[
      ylabel={\small{LF1 (100\%)}},
      legend style={
        font=\scriptsize,
        cells={anchor=west}
      },
      legend pos=north west,
      symbolic x coords={1-10,11-20,21-30,31-40,41-50,51-60},
      xtick=data,
      ybar,
    ]
    \addplot coordinates {(1-10, 96.02)(11-20, 95.60)  (21-30, 95.49)  (31-40, 94.27)
 (41-50, 93.78)  (51-60, 95.93)};
    \addplot coordinates {(1-10, 95.12)(11-20, 95.64)  (21-30, 95.18)  (31-40, 94.16)
 (41-50, 93.32)  (51-60, 96.23)};
  \end{axis}
 \end{tikzpicture}
 }%
 
\caption{DM (ID)}
\label{error_a}
\end{subfigure}
	\begin{subfigure}[t]{0.32\linewidth}
		 \resizebox{\textwidth}{!}{%
		 
		 \begin{tikzpicture}
  \begin{axis}[
      ylabel={\small{LF1 (100\%)}},
      legend style={
        cells={anchor=west}
      },
      legend pos=north east,
      symbolic x coords={1-10,11-20,21-30,31-40,41-50,51-60},
      xtick=data,
      ybar,
    ]
    \addplot coordinates {(1-10, 97.53)(11-20, 96.60)  (21-30, 95.57)  (31-40, 94.23)
 (41-50, 95.05)  (51-60, 94.50)};
    \addplot coordinates {(1-10, 96.23)(11-20, 95.81)  (21-30, 94.79)  (31-40, 93.65)
 (41-50, 94.13)  (51-60, 95.32)};
  \legend{Ours, w/o label corr.}
  \end{axis}
 \end{tikzpicture}
}
\caption{PAS (ID)}
\label{error_a}
\end{subfigure}
	\begin{subfigure}[t]{0.32\linewidth}
		 \resizebox{\textwidth}{!}{%
		 
		 \begin{tikzpicture}
  \begin{axis}[
      ylabel={\small{LF1 (100\%)}},
      legend style={
        font=\scriptsize,
        cells={anchor=west}
      },
      legend pos=north west,
      symbolic x coords={1-10,11-20,21-30,31-40,41-50,51-60},
      xtick=data,
      ybar,
    ]
    \addplot coordinates {(1-10, 87.22)(11-20, 85.07)  (21-30, 82.72)  (31-40, 81.26)
 (41-50, 81.53)  (51-60, 84.22)};
    \addplot coordinates {(1-10, 87.90)(11-20, 85.10)  (21-30, 82.54)  (31-40, 80.86)
 (41-50, 80.52)  (51-60, 82.17)};
  \end{axis}
 \end{tikzpicture}
 }%
 
\caption{PSD (ID)}
\label{error_a}
\end{subfigure}

	\begin{subfigure}[t]{0.32\linewidth}
		 \resizebox{\textwidth}{!}{%
		 
		 \begin{tikzpicture}
  \begin{axis}[
      ylabel={\small{LF1 (100\%)}},
      legend style={
        font=\scriptsize,
        cells={anchor=west}
      },
      legend pos=north east,
      symbolic x coords={1-10,11-20,21-30,31-40,41-50,51-60,61-70},
      xtick=data,
      ybar,
    ]
    \addplot coordinates {(1-10, 93.42)(11-20, 92.97)  (21-30, 91.65)  (31-40, 90.49)
 (41-50, 89.45)  (51-60, 87.84) (61-70, 84.98)};
    \addplot coordinates {(1-10, 93.56)(11-20, 92.98)  (21-30, 91.63)  (31-40, 90.11)
 (41-50, 88.95)  (51-60, 86.93) (61-70, 85.80)};
  \end{axis}
 \end{tikzpicture}
 }%
\caption{DM (OOD)}
\label{error_a}
\end{subfigure}
	\begin{subfigure}[t]{0.32\linewidth}
		 \resizebox{\textwidth}{!}{%
		 
		 \begin{tikzpicture}
  \begin{axis}[
      ylabel={\small{LF1 (100\%)}},
      legend style={
        font=\scriptsize,
        cells={anchor=west}
      },
      legend pos=north west,
      symbolic x coords={1-10,11-20,21-30,31-40,41-50,51-60,61-70},
      xtick=data,
      ybar,
    ]
    \addplot coordinates {(1-10, 94.01)(11-20, 94.30)  (21-30, 93.38)  (31-40, 92.61)
 (41-50, 91.67)  (51-60, 90.48) (61-70, 94.73)};
    \addplot coordinates {(1-10, 93.93)(11-20, 93.80)  (21-30, 92.73)  (31-40, 92.04)
 (41-50, 90.72)  (51-60, 90.65) (61-70, 94.93)};
  \end{axis}
 \end{tikzpicture}
 }%
\caption{PAS (OOD)}
\label{error_a}
\end{subfigure}	\begin{subfigure}[t]{0.32\linewidth}
		 \resizebox{\textwidth}{!}{%
		 
		 \begin{tikzpicture}
  \begin{axis}[
      ylabel={\small{LF1 (100\%)}},
      legend style={
        font=\scriptsize,
        cells={anchor=west}
      },
      legend pos=north west,
      symbolic x coords={1-10,11-20,21-30,31-40,41-50,51-60,61-70},
      xtick=data,
      ybar,
    ]
    \addplot coordinates {(1-10, 89.20)(11-20, 83.99)  (21-30, 80.66)  (31-40, 79.11)
 (41-50, 79.40)  (51-60, 74.24) (61-70, 77.02)};
    \addplot coordinates {(1-10, 89.40)(11-20, 84.40)  (21-30, 79.84)  (31-40, 77.94)
 (41-50, 77.45)  (51-60, 70.43) (61-70, 77.63)};
  \end{axis}
 \end{tikzpicture}
 }%
\caption{PSD (OOD)}
\label{error_a}
\end{subfigure}
\caption{LF1 of different sentence lengths on three semantic formalisms.}
\label{fig:error}
\end{figure*}

Table \ref{tab:sdp_result} shows the results on three benchmarks of semantic dependency parsing. The work of \citet{DBLP:conf/flairs/HeC20} is reported for reference, since they used Flair contextualized embedding \cite{akbik2018coling} in addition while we did not.  Our main baseline is {\it Pointer} \cite{fernandez-gonzalez-gomez-rodriguez-2020-transition}, which leverages pointer networks and has the same setting (we both used the base version of BERT \cite{devlin-etal-2019-bert} and lemma embeddings) as ours. In our preliminary experiments, we find character-level word embedding has no benefit, thus we discard it for simplicity.
As we can see from the table, our model outperforms Pointer by 0.3 and 0.4 LF1 in ID and OOD test sets in average. Notably, our model achieves the best performance on ID test set of DM,  OOD test set of DM and PSD among all listed models.
 

\section{Analysis}
\subsection{Ablation study}
In Table \ref{tab:sdp_result}, ``w/o label correlation'' amounts to the mean-field model of \citet{wang-etal-2019-second}, but uses the same neural encoder and hyper-parameters as ours for fair comparison. As we can see, modeling label correlations brings 0.6 and 0.4 average LF1 score improvement in ID and OOD (test sets) respectively, validating its effectiveness.  The improvement on DM is the smallest: only 0.2 and 0.1 LF1 score in ID and OOD. We speculate that this is because DM has the most coarse-grained labels, as we can see from Fig.\ref{fig:dep} (b). PAS has a more fine-grained label set (Fig.\ref{fig:dep}(b) vs. Fig.\ref{fig:dep}(c) for example), and we observe a larger improvement: 0.8 and 0.5 LF1 in ID and OOD. Previous works found that PSD is the most difficult to learn as it has the most fine-grained label set and the largest label set size \cite{wang-etal-2019-second}, and modeling label correlations results in 0.3 and 0.6 LF1 improvement in ID and OOD. This ablation study indicates that it is more advantageous to model label correlations when the label set size is large and the labels are fine-grained.

\subsection{Error analysis}
Fig.\ref{fig:error} plots the LF1 against different sentence lengths in three semantic formalisms. We compare our model with ``w/o label correlation''. 
We can see that modeling label correlations benefit the prediction of sentences of medium length (21-50). When the sentence length is small (1-20) or large (51-70), modeling label correlations  has no clear advantages.

\subsection{Speed comparison}
\begin{figure}[h!]
\centering
\scalebox{0.8}{
\begin{tikzpicture}
\begin{axis}[
    symbolic x coords={1, 5, 10,20,30,40},
    xtick=data,
    xlabel=Label set size,
    ylabel= time (s)]
    \addplot coordinates {
        (1, 0.44)
        (5, 0.87)
        (10, 1.86)
        (20, 3.49)
        (30, 5.88)
        (40, 9.73)
    };
        \addlegendentry{w/o CPD}

        \addplot[mark=o,mark options={solid},red,thick,dashed] coordinates {
        (1, 1.55)
        (5, 1.60)
        (10, 1.61)
        (20, 1.70)
        (30, 1.75)
        (40, 1.85)
    };
    \addlegendentry{w/ CPD}
\end{axis}
\end{tikzpicture}
}
\caption{Total running time.}
\label{compare_time}
\end{figure}
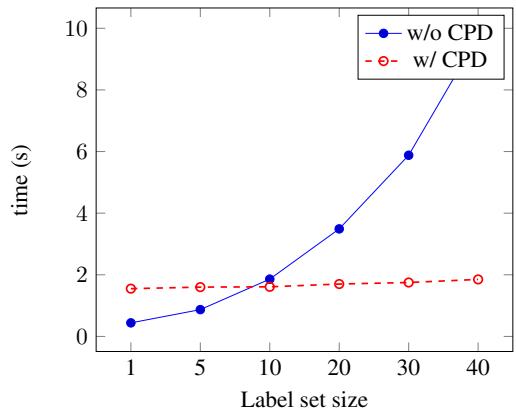

To show the advantage of applying CPD to accelerate labeled mean-field inference, we compare the running speed of two cases: (1) apply CPD (``w/ CPD'') (2) not apply CPD (``w/o CPD''). The first case is what we described previously. In the latter case, we first recover the order-5 second-order score tensors, then run the standard mean-field inference (i.e., the not accelerated one).  We set the rank size to 300, batch size to 1, the number of mean-field inference iteration to 3, and report the total time of running 100 times. The experiment is conducted on a single Titan V GPU.   We plot the change of running time with the change of label set size $L$ in Fig. \ref{compare_time}.  We can see that when $L=1$, i.e., unlabeled mean-field inference, ``w/ CPD'' is slower. Although it has a quadratic complexity, which is lower than the cubic complexity of ``w/o CPD'', it introduces a constant that is larger than the sentence length. When increasing the label size, the running time of ``w/o CPD'' grows quadratically, and becomes much slower than ``w/ CPD''. Notably, it is common to have a label set size $> 50$, e.g., PSD has 91 labels, making ``w/o CPD'' impractical to use. Besides, ``w/o CPD'' quickly encounters the out-of-memory issue due to the large space complexity. On the other hand, ``w/ CPD'' is much more memory efficient, as the second-order score tensors have no need to be materialized.

\section{Related work}
\paragraph{Semantic dependency parsing.}  Since the SemEval 2015 Task 18 \cite{oepen-etal-2015-semeval}, there are many studies in semantic dependency parsing, which can mainly be categorized into two groups:  graph-based methods and transition-based methods. For graph-based methods,
\cite{almeida-martins-2015-lisbon} adapt the Turbo parser \cite{martins-etal-2013-turning}, which is originally designed for syntactic dependency parsing, to produce DAGs. It additionally leverages co-parent information \cite{martins-almeida-2014-priberam}, and adopts AD$^3$ algorithm for approximate decoding. This work is extended by \cite{peng-etal-2017-deep} with BiLSTM feature extraction and multi-task learning.  \citet{sun-etal-2017-semantic} propose a Maximum Subgraph parsing algorithm and \citet{chen-etal-2018-neural} conduct experiments using this algorithm. \citet{cao-etal-2017-quasi} devise a novel algorithm for quasi-second-order Maximum Subgraph parsing.  \citet{dozat-manning-2018-simpler} adapt the seminal Biaffine Parser \cite{DBLP:conf/iclr/DozatM17} to perform semantic dependency parsing. This work is extended by \citet{wang-etal-2019-second} who introduce a similar deep Triaffine attention to score second-order factors, and unroll mean-field inference or belief propagation for end-to-end learning; and by \citet{DBLP:conf/flairs/HeC20} who use contextual string embeddings (i.e., Flair \cite{akbik2018coling}) to enhance performance.   \citet{jia-etal-2020-semi} use CRF-autoencoder \cite{DBLP:conf/nips/AmmarDS14} for semi-supervised semantic dependency parsing. 

As for transition-based methods, \citet{DBLP:conf/aaai/WangCGL18} adapt the list-based arc-eager transition system \cite{choi-mccallum-2013-transition} for neural semantic dependency parsing. \citet{kurita-sogaard-2019-multi} use reinforcement learning to build DAGs sequentially.  \citet{fernandez-gonzalez-gomez-rodriguez-2020-transition} adapt the left-to-right dependency parser of \citet{fernandez-gonzalez-gomez-rodriguez-2019-left} to produce DAGs.

\paragraph{Higher-order syntactic dependency parsing.} Considering that there are many works adapting higher-order syntactic parsers for semantic dependency parsing, it is worthy to introduce some related works of higher-order syntactic parsing.   
Before the deep learning era, first-order dependency parsing \cite{mcdonald-etal-2005-online} has been considered insufficient in capturing rich contextual information. To capture higher-order information, researchers develop many interesting dynamic programming algorithms for higher-order parsing \cite{mcdonald-pereira-2006-online, carreras-2007-experiments, koo-collins-2010-efficient, ma-zhao-2012-fourth}. However, these algorithms have two drawbacks: (1) they can only handle projective trees as higher-order nonprojective parsing is NP-hard \cite{mcdonald-pereira-2006-online} and (2) they have a high parsing complexity when the order is greater than two. For instance, third-order parsing has an $O(n^4)$ parsing complexity \cite{koo-collins-2010-efficient}.  The parsing community then resorts to approximate algorithms to tackle the aforementioned problems. \citet{smith-eisner-2008-dependency} use loopy belief propagation to do third-order projective parsing in cubic time without much accuracy degeneration. \citet{koo-etal-2010-dual,rush-etal-2010-dual} use dual decomposition for higher-order nonprojective dependency parsing. However, the subgradient algorithm of dual decomposition is inefficient when there are many overlapping components \cite{martins-etal-2011-dual}. 
To tackle this, \citet{martins-etal-2011-dual, martins-etal-2013-turning} use alternating directions dual decomposition (AD$^3$ \cite{DBLP:conf/icml/MartinsFASX11, DBLP:journals/jmlr/MartinsFASX15}) instead for faster convergence. In the deep learning age, researchers use neural networks for scoring instead of relying on hand-crafted features. 
Inspired by the deep biaffine attention \cite{DBLP:conf/iclr/DozatM17}, \citet{wang-tu-2020-second,zhang-etal-2020-efficient} use the deep triaffine attention to score second-order factors for their second-order parsers, and show better performance than the first-order biaffine parser. \citet{wang-tu-2020-second} use mean-field inference and \citet{fonseca-martins-2020-revisiting} use AD$^3$ for second-order nonprojective dependency parsing.

\paragraph{Graphical models with approximate inference in NLP.} Many structured prediction tasks in NLP can be formulated as  graphical model inference, such as dependency parsing \cite{smith-eisner-2008-dependency}, constituency parsing \cite{naradowsky-etal-2012-grammarless}, named entity recognition \cite{durrett-klein-2014-joint, naradowsky-etal-2012-grammarless},  semantic role labeling \cite{naradowsky-etal-2012-improving}, coreference resolution \cite{durrett-klein-2014-joint}, among others. They use loopy belief propagation or mean-field inference for approximate decoding. In the deep learning age, end-to-end mean-field inference becomes very popular and has been used in various NLP tasks, such as dependency parsing \cite{wang-etal-2019-second,wang-tu-2020-second}, semantic role labeling \cite{li-etal-2020-high, DBLP:journals/corr/abs-2112-02970, Jia2022Span}, structured sentiment analysis \cite{DBLP:journals/corr/abs-2109-06719} etc.

\paragraph{Accelerate parsing with CPD.} 
CPD has been previously used to accelerate parsing algorithms in the parsing literature.  \citet{cohen-etal-2013-approximate, DBLP:conf/nips/CohenC12} use CPD to accelerate (latent-variable) PCFG parsing. \citet{yang-etal-2021-pcfgs, yang-etal-2021-neural} use CPD to accelerate the inside algorithms of (lexicalized) PCFGs. In this work, we leverage CPD to accelerate second-order semantic dependency parsing with mean-field inference. Our work is also closely related to \cite{DBLP:journals/corr/abs-2010-09283} who use CPD to accelerate higher-order belief propagation.

\section{Conclusion}
In this work, we have presented a simple and efficient method to model label correlations for second-order semantic dependency parsing. We leveraged CP decomposition to decrease the computational complexity of mean-field inference from cubic to quadratic. Experiments on SemEval 2015 Task 18 English datasets validated the effectiveness of modeling label correlations.

\bibliography{anthology,custom}
\bibliographystyle{acl_natbib}

\appendix

\end{document}